\documentclass{sig-alternate-05-2015}
\usepackage{tikz}
\usepackage{amsmath}
\usetikzlibrary{fit,positioning}
\usepackage[]{algorithm2e}
\usepackage{booktabs}
\usepackage{balance}
\usepackage{tabularx}
\usepackage{array}
\newcolumntype{L}[1]{>{\raggedright\let\newline\\\arraybackslash\hspace{0pt}}m{#1}}
\newcolumntype{C}[1]{>{\centering\let\newline\\\arraybackslash\hspace{0pt}}m{#1}}
\newcolumntype{R}[1]{>{\raggedleft\let\newline\\\arraybackslash\hspace{0pt}}m{#1}}
\usepackage{bm}

\begin{document}

\setcopyright{acmcopyright}



\conferenceinfo{Neu-IR '16 SIGIR Workshop on Neural Information Retrieval}{July 17--21, 2016, Pisa, Italy}

%

\title{An empirical study on large scale text classification with skip-gram embeddings}

%
%
%
%
%

\numberofauthors{2} 
%
\author{
%
%
\alignauthor
Georgios Balikas\\
       \affaddr{University of Grenoble Alpes/Coffreo}\\
       \affaddr{Grenoble, France}\\
       \email{Georgios.Balikas@imag.fr}
\alignauthor
Massih-Reza Amini\\
       \affaddr{University of Grenoble Alpes}\\
       \affaddr{Grenoble, France}\\
       \email{Massih-Reza.Amini@imag.fr}
}
\date{30 July 1999}

\maketitle
\begin{abstract}
We investigate the integration of word embeddings as classification features in the setting of large scale text classification. Such representations have been used in a plethora of tasks, however their application in classification scenarios with thousands of classes has not been extensively researched, partially due to hardware limitations.
In this work, we examine efficient composition functions to obtain document-level from word-level embeddings and we subsequently investigate their combination with the traditional one-hot-encoding representations.
By presenting empirical evidence on large, multi-class, multi-label classification problems, we demonstrate the efficiency and 
the performance benefits of this combination.
\end{abstract}

%
%
%
%
%

%
%


\section{Introduction and Preliminaries}

With the proliferation of text data available online,
text classification has attracted a lot of interest. 
Traditionally, $N$-grams are considered as
document features and are subsequently fed to a classifier
such as Support Vector Machines (SVMs) \cite{cortes1995support,joachims1999making}. One-hot-encoding representations, although prominent in the literature, have two significant drawbacks: (i) they result in a very high dimensional and sparse feature space, and (ii) they do not encode similarity between words. Lately, a lot of research has been devoted to the direction of distributed representations \cite{harris1954distributional}. Distributed representations of words, are continuous, low dimensional, dense vectors that characterize the meaning and the semantic content of words. Each dimension
of the embedding represents a latent feature of the
word, hopefully capturing useful syntactic and semantic
properties. As a result, semantically similar words, such as ``strong'' and ``powerful'', will be close in the output vectorial space.

In this work, we present a focused contribution as part of an interesting classification application. We investigate the performance of word embeddings learned using the skip-gram model \cite{mikolov2013efficient} in the context of large scale, multi-label document classification. We report results on real-world classification problems with up to 10,000 classes and we demonstrate a straightforward way to combine document-level embeddings with one-hot-encoding representations. The contributions of our work are twofold: we, first, propose an efficient way to achieve satisfactory classification performance using only embedding representations and,  we further improve it by applying a naturally parallelizable fusion mechanism between the distributed representations and one-hot-encoding representations.

Several methods have been proposed for obtaining distributed word-level representations. We cite for instance:  \cite{mikolov2013distributed,mikolov2013efficient,collobert2011natural,levy2014dependency,pennington2014glove}. Extensions of those methods have been proposed to cope with larger portions of text such as sentences or paragraphs: \cite{zhao2015self,le2014distributed,socher2011dynamic}. However, generalizing from words to larger text spans can be inefficient, since for every unseen span new passes over a neural network are required. Hence, we focus on methods that given a dictionary of word representations apply composition functions \cite{mitchell2010composition,blacoe2012comparison} to produce document representations. 

Although distributed embeddings have been applied in a plethora of tasks from analogies evaluation \cite{levy2014linguistic} to extractive summarization \cite{kaageback2014extractive}, their application on large scale text classification has not been investigated. What is more, most of the work in this direction deals with short text spans, such as sentences or tweets, and the size of the investigated problems with regard to the number of classes is small. Recently, for instance, \cite{kenter2015short,kim2015convolutional,song2015unsupervised} investigated the problem of short text similarity with applications to classification with a limited number of classes, like binary sentiment analysis \cite{nakov2013semeval} and ternary sentiment analysis \cite{pang2005seeing}. More frequently, word embeddings for text classification are used for initializing architectures such as convolutional  and recurrent networks. The works of \cite{kalchbrenner2014convolutional} and \cite{kim2014convolutional} for instance, are in this line but, again, they limit their study at sentence-length spans. The latter is, also, due to hardware limitations of the GPUs used: their limited memory capacity in conjunction with the vocabulary size of large scale classification problems require many data transfer operations which results in significant overhead. 
In this work we place ourselves at the large scale setting (number of classes in the order of $10^4$) where compositional methods based on neural networks are difficult to be applied.


The remainder of this paper is organised as follows: in Section 2 we propose and evaluate composition functions for obtaining document from word representations, and in Section 3 we discuss their integration with one-hot-encoding schemes. Finally, Section 4 concludes  with remarks to our future work.

\section{Distributed Representations as Classification Representations}
\setlength{\tabcolsep}{2pt} 
\begin{table}\scriptsize\centering
 \begin{tabular}{lccC{2cm}C{1.8cm}}

  \toprule
  & Instances & Vocabulary & Avg. labels per instance &  Avg. Doc. Size (words) \\
  \midrule
  Train Data & 12.5M& 868,219 & - & 215 \\
  PubMed$_{1000}$ &225,000&289,789&1.42&239\\
  PubMed$_{5000}$ &225,000&302,802&3.23&233\\
  PubMed$_{10000}$ &225,000&303,637&6.64&230\\
  \bottomrule
 \end{tabular}\caption{Description of the data used to obtain the distributed representations (Train Data) and for classification purposes.}\label{table:data}
\end{table}

We explore three functions to compose document representations: \textit{min}, \textit{average} and \textit{max} 
\cite{collobert2011natural,socher2011dynamic} which have been used as simple yet
effective methods for compositionality learning
in vector-based semantics \cite{mitchell2010composition}, to obtain a document's representation. 
We use the output of each composition function as document features and we evaluate both their classification performance as well as the performance of their concatenation:
\[
 z_{conc}(\text{doc}) = [z_{avg}(doc), z_{min}(doc), z_{max}(doc)]
 \]
where $z(doc)$ is the representation of a document and $z_x(doc)$ is the result of applying the composition $x\in\{avg, max, min\}$ element-wise, in the distributed vectors of the words of the documents. For instance, assuming we want the representation of ``harsh winter'', the composition function requires the vector representations of the words ``harsh'' and ``winter''. Then, applying  the \textit{max} function will produce a vector of the same dimensionality with the original representations where each element will have the maximum value of the corresponding elements of the original word representations. In this process, we assume access to a vocabulary of distributed representations, where each word is associated with a $D$-dimensional vector. This vocabulary, which we assume readily available, may have been generated beforehand.

%

\begin{figure*}\centering
\includegraphics[scale=0.47]{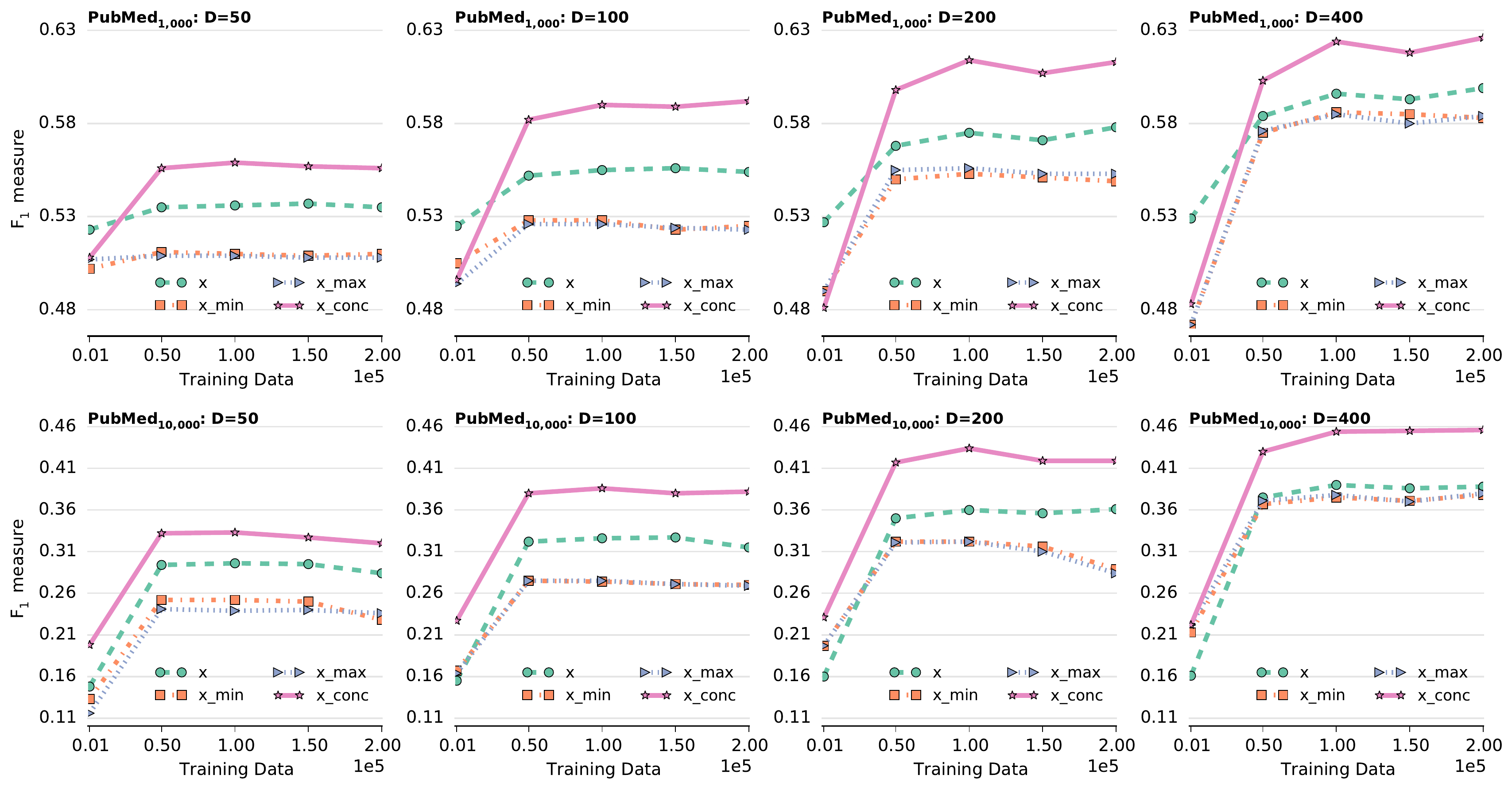}
\caption{Classification performance of the different composition functions on the PubMed$_{1000}$ and PubMed$_{10000}$ datasets with representation dimensions $D\in\{50,100,200,400\}$.}\label{fig:class1}
\end{figure*}

Table \ref{table:data} shows the data we use throughout the paper. They are abstracts of biomedical texts from PubMed released by the BioASQ challenge organisers \cite{tsatsaronis2015overview} as well as texts from Wikipedia \cite{partalas2015lshtc}.
To obtain the  dictionary of word embeddings we used the skipgram model of \texttt{word2vec} tool \cite{mikolov2013efficient}.  We have kept the tool's default parameters for skip-gram apart from  the number of iterations that we have set to 15.
For training the representations, we used 10M PubMed abstracts and we added 2.5M Wikipedia documents for better generalization (``Train Data'' of Table \ref{table:data}). 
In the preprocessing steps we applied lower-casing, we space-padded the punctuation symbols and we filtered the words with less than  5 occurrences.  
For classification purposes we used the remaining PubMed abstracts from the released data. We created three classification datasets: ``PubMed$_x$'' with $x\in\{1,000,5,000, 10,000\}$ being  the $x$ most common classes in the data. We performed a stratified split in train-test (200K-25K instances) parts. 
In the experiments hereafter, we use SVMs with linear kernel which have been widely used for text classification. The $\lambda$ value that controls the importance of the regularization term in the optimization problem was selected using 5-fold cross validation on the training data of each experiment and for each representation.  The classification problem is multi-label: each instance is associated with several classes as shown in  Table \ref{table:data}. We cope with the multi-label problem using a binary relevance approach \cite{tsoumakas2006multi}. For SVMs, tackling the multi-label problem with binary relevance results in predicting for each instance every label with positive distance from the separating hyperplanes of the one-vs-rest binary problems. If the classifier does not return any label for an instance, the most common label of the training data is assigned. For our implementations, we have used Python's Scikit-Learn \cite{scikit-learn}. 

We now present results for classification experiments on the PubMed$_{1,000}$ and PubMed$_{10,000}$ datasets when the described composition functions are used.  Figure \ref{fig:class1} shows the scores for the $F_1$ measure obtained on the test data, when varying the  size of the training data.
From the Figure, first note that the \textit{avg} function performs better compared to both \textit{min} and \textit{max}. The latter two, perform equally but they are not competitive. However, the best results are consistently obtained with the concatenation (\textit{conc}) of the outputs of the three composition functions.   Interestingly, adding the \textit{min} and \textit{max} representations creates a richer representation that benefits the performance. To this direction, note the steep increase in the performance of \textit{conc} representations with the availability of training data: being richer, those representations have bigger discriminative power. Depending on the dataset and the dimension of the representations, the achieved improvements using \textit{conc} vary from $\sim$3-5  $F_1$ points  which is important for such problems. We believe that the \textit{avg} function does not retain enough information for large documents, given that they  consist on average of more than 200 words (Table \ref{table:data}). To this end, Table \ref{table:docDimensionImpact} reports the micro $F_1$ measure for the PubMed$_{10,000}$ dataset, with respect to the document length in words. Note that the best performance is achieved for smaller documents independently of the embedding dimension.

\setlength{\tabcolsep}{4pt} 
\begin{table}\scriptsize\centering
 \begin{tabular}{lccccc}

  \toprule
  &\multicolumn{5}{c}{Document length (in words)}\\
  \cmidrule(lr){2-6}
   & <100& 100-200 & 200-300 &300-400& >400  \\
  \midrule
  D=50 &0.312&0.268&0.266&0.282 & 0.307\\
  D=200 & 0.365,& 0.339& 0.337& 0.345& 0.359\\
  D=400 &0.427&0.400&0.396&0.404 & 0.409\\
  \bottomrule
 \end{tabular}\caption{The impact of the document length on the classification performance (micro $F_1$ measure) when using \textit{avg} representations and the embeddings dimension is $D\in\{50,200,400\}$. The representations perform better on smaller documents, which is in line with the outcome that the \textit{avg} representations do not capture enough information for large documents.}\label{table:docDimensionImpact}
\end{table}

Another observation from Figure \ref{fig:class1} and Table \ref{table:docDimensionImpact} concerns the effect of the dimension of word representations in the classification performance. Representations of bigger dimensions benefit performance. In fact, increasing the dimension from 50 to 400, improves the $F_1$ measure for \textit{conc} for PubMed$_{1,000}$ (resp. PubMed$_{10,000}$) by $\sim$7 (resp. $\sim$13) points. 
Summarizing, we highlight here two of the advantages of the proposed approach: (i) although simple, our composition functions have yielded significant performance improvements and, (ii) their application on large datasets is naturally parallelizable, hence, they can be easily applied on real, large scale problems.


\setlength{\tabcolsep}{6pt} 
\begin{table*}[t]\small\centering
 \begin{tabular}{lccccccccc}\toprule
 &\multicolumn{3}{c}{PubMed$_{1,000}$} &\multicolumn{3}{c}{PubMed$_{5,000}$} &\multicolumn{3}{c}{PubMed$_{10,000}$} \\
 \cmidrule(lr){2-10}
 \textit{hash}&\multicolumn{3}{c}{0.63} &\multicolumn{3}{c}{0.427}&\multicolumn{3}{c}{0.456}\\
\textit{tf-idf}&\multicolumn{3}{c}{0.65} &\multicolumn{3}{c}{0.469}&\multicolumn{3}{c}{0.492}\\\cmidrule(lr){1-10}

 &D=100&D=200&D=400&D=100&D=200&D=400&D=100&D=200&D=400\\
  \cmidrule(lr){2-4} \cmidrule(lr){5-7}\cmidrule(lr){8-10}

\textit{x\_conc}&0.592 &0.614&0.626&0.362&0.41&0.436&0.386&0.434&0.454\\
\textit{hash+x\_conc}&0.651 &0.654&0.646&0.464&0.476&0.473&0.488&0.495&0.491\\
\textit{tfidf+x\_conc}&\textbf{0.66}$^\dagger$  &\textbf{0.66}$^\dagger$ &0.656&0.484$^\dagger$ &0.486$^\dagger$ &\textbf{0.487}$^\dagger$ &0.507$^\dagger$ &0.507$^\dagger$ &\textbf{0.512}$^\dagger$\\\bottomrule
 \end{tabular}\caption{Classification performance of the different representations. The upper part of the table presents one-hot-encoding methods, while the bottom part methods that depend on the dimensionality of the distributed representations. We report the best performance obtained when the size $N$ of the training data $N\in\{1,50,100, 150, 200\} \times 10^3$. The best achieved performance per classification problem is shown in bold. We have performed statistical significance tests to test if improvements obtained compared to \textit{tf-idf} representations are statistically important, which is noted by putting a dagger ($\dagger$) to the accuracy scores.}\label{table:classification}
\end{table*}

\section{Combination of Distributed and One-Hot-Encoding Schemes}
\begin{figure}\centering
 \includegraphics[width=0.45\textwidth]{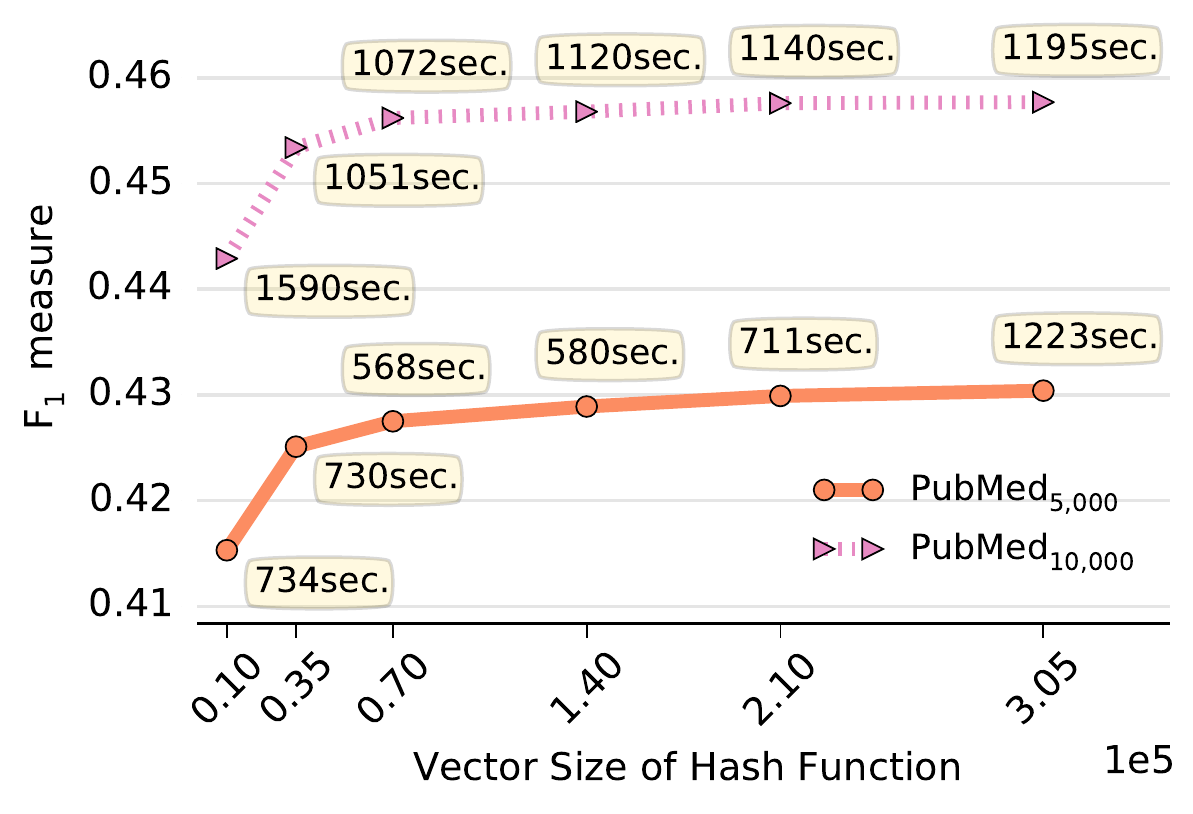}
 \caption{The effect of the vector size generated by the hash function in terms of classification performance and time efficiency for PubMed$_{5,000}$ and PubMed$_{10,000}$.}\label{fig:hashDim}
\end{figure}

In the previous section we have shown that the concatenated representations obtained by the output of the composition functions achieved the best classification performance.  
We, now, compare them with the traditional one-hot-encoding representations. We focus on two ways of representing text: (i) by using \textit{tf-idf} representations of unigrams, and (ii) by employing a \textit{hash} function \cite{forman2008extremely,sutton1998reinforcement}. 
For the former, to generate the tf-idf representations we used sub-linear term frequency counts multiplied by their respective smoothed inverse document frequency, i.e., for a term $w$ we have $(1+\log (tf_w))*(idf_w+1)$.
For the latter, the hash function, given a text string, transforms it on a numerical value in a pre-specified space, that is used as the index to generate a vector representation. Increasing the output dimension of the hash functions reduces the probability of collisions, i.e. different words mapped on the same vector indices,  but also increases the output vector size. We investigate this trade-off in Figure \ref{fig:hashDim}: we present the effect of the size of the hash representations with respect to classification performance for PubMed$_{5,000}$ and PubMed$_{10,000}$ which have the biggest vocabulary size. We also report exact timings of each scenario, executed on 10 cores of an Intel(R) Xeon(R) CPU E5-2640 v3 @ 2.60GHz. From the figure, note that after 70K features the classification performance does not improve when increasing the size of the feature space. On the other hand, increasing the representation's dimension, the training time also increases. For reference, \textit{tf-idf} representations in the same computational setting need 602 sec. and 1203 sec. for PubMed$_{5,000}$ and PubMed$_{10,000}$ respectively. As a result, we set the hash dimension to 70,000 features. Note the  significant dimensionality reduction achieved, given the vocabulary sizes of our problems reported in Table \ref{table:data}.

Table \ref{table:classification} details the scores achieved with regard to the representations used. We first discuss the performance when single representations are used: \textit{x\_conc}, \textit{tf-idf} and \textit{hash}.  Notice that \textit{tf-idf} performs better than both \textit{x\_conc} and \textit{hash}, with the latter achieving the lowest performance. The performance of the three representations on PubMed$_{1,000}$ is comparable, but in the bigger classification problems \textit{tf-idf} and performs considerably better. We, thus, consider the \textit{tf-idf} representations as our baseline model, and we examine how the models with concatenated representations behave compared to it. 

We investigate now whether the fusion of distributed document representations with the one-hot-encoding representations benefits the classification performance. 
The last two lines of Table \ref{table:classification} present the performance of the fusion, by concatenation, of the embedding representations with $D\in\{100,200,400\}$ with \textit{hash} and \textit{tf-idf}. In the experiments, both \textit{hash} and \textit{tf-idf} consistently achieve better performance when combined with the distributed representations. 
For instance, for PubMed$_{10,000}$ and $D=400$ the \textit{tf-idf} (resp. \textit{hash}) representations improve in absolute numbers by 2 (resp. 3.5)  $F_1$ points. We have performed two-sided student's t-tests ($p<0.01$) to compare whether the improvements obtained for each classification problem are statistically significant compared to using \textit{tf-idf} representations. Those results, indicated by $(^\dagger)$ in Table \ref{table:classification}, reveal that the concatenation of \textit{tf-idf} with distributed representations improves the classification performance in a statistically significant way. 
In addition to the important improvements, note than in the fused representations, the effect of $D$ diminishes. For instance in PubMed$_{1,000}$, one can obtain the optimal performance using embedding dimensions with $D<400$ and similar observations can be made in the rest of the datasets.     

\section{Conclusion and Discussion}
In this work we have restricted ourselves in representations learned in the word level. This is advantageous in terms of speed since the dictionary of representations can be generated offline. Then, applying composition functions is naturally parallelizable and fast for prediction. However, this poses the challenge of having robust composition functions, which if carefully selected, can result in performance gains such as those reported above. Also, similarly to the bag-of-words paradigm, it does not take into account the word order and the words' grouping in coherent segments like sentences or phrases.

In this line, it would be interesting to further investigate how more complex embeddings such as paragraph vectors \cite{le2014distributed} perform. Even if such approaches are computationally  expensive in the document level, previous research has shown their effectiveness on the sentence level. Hence, a direct extension of this work is to test the investigated composition functions with  sentence level representations. In terms of applications, those sentence representations can be directly used to evaluate the effectiveness of the embeddings and the composition functions. Importantly, their can be evaluated simultaneously in different levels of text granularity from sentences to documents in the framework of passage retrieval and document classification from instance. 

Another interesting line of research concerns the memory efficiency of such dense representations. Recent research efforts \cite{balikas2016multi,jurgovsky2016evaluating} have investigated ways of compressing the learned (or composed) representations using either linear (e.g. PCA) or non-linear (e.g. auto-encoders) approaches to decrease the memory requirements or the dimension of the representations. Such approaches, apart from having a positive effect on the memory footprint, also have a positive effect on the required computational requirements for training.

In this work we have evaluated different composition functions for obtaining document-level representations using distributed embeddings of words. Summarizing our findings, adding the concatenated vector of word-level skip-gram derived features to tf-idf unigrams performs better
than tf-idf features alone. Also, the result seems to be more pronounced as the representations' cardinality increases and as the output label space increases. Given the obtained improvements, we have also outlined promising future research directions.

\balance

\bibliographystyle{abbrv}

%
%
\end{document}